\documentclass[10pt,conference]{IEEEtran}
\IEEEoverridecommandlockouts

\usepackage{cite}
\usepackage{amsmath,amssymb,amsfonts}
\usepackage{graphicx}
\usepackage{textcomp}
\usepackage{xcolor}
\usepackage{booktabs}
\usepackage{array}
\usepackage{url}
\usepackage{tikz}
\usetikzlibrary{positioning,arrows.meta,fit}

\newcommand{\pass}{\textsc{pass}}
\newcommand{\fail}{\textsc{fail}}
\newcommand{\skipv}{\textsc{skip}}

\begin{document}
\bstctlcite{IEEEexample:BSTcontrol}

\title{AutoRefine: Auto-Calibrating Code Judges\\from Workflow Outcomes}

\author{\IEEEauthorblockN{Shervin Shahidi\textsuperscript{*}}
\IEEEauthorblockA{\textit{Meta Platforms}\\
shervin@meta.com
}
\and
\IEEEauthorblockN{James Poon\textsuperscript{*}}
\IEEEauthorblockA{\textit{Meta Platforms}\\
jypoon@meta.com
}
\and
\IEEEauthorblockN{Hang Xie}
\IEEEauthorblockA{\textit{Meta Platforms}\\
hangxie@meta.com
}
\and
\IEEEauthorblockN{Kenneth O. Xu}
\IEEEauthorblockA{\textit{Meta Platforms}\\
kennethoxu@meta.com
}
\and
\IEEEauthorblockN{Robin H{\aa}kanson}
\IEEEauthorblockA{\textit{Meta Platforms}\\
matshakanson@meta.com
}
\and
\IEEEauthorblockN{Vikram Nair}
\IEEEauthorblockA{\textit{Meta Platforms}\\
vikramnair@meta.com
}
\and
\IEEEauthorblockN{Jens Pillgram-Larsen}
\IEEEauthorblockA{\textit{Meta Platforms}\\
jenspl@meta.com
}
\and
\IEEEauthorblockN{Nachiappan Nagappan}
\IEEEauthorblockA{\textit{Meta Platforms}\\
nnachi@meta.com
}
\thanks{\textsuperscript{*}These authors contributed equally to this work}}

\maketitle
\begin{abstract}
LLM judges increasingly serve as quality gates for code changes that have no test oracle, but a judge prompt must be
calibrated for its task and recalibrated as the system it judges evolves, while hand labels for the calibration are resource intensive and go stale. Development workflows emit outcome labels for free (tests pass or fail, changes land or are abandoned), but
these labels are noisy. We present AutoRefine, a loop that rewrites a deployable judge's prompt to reproduce outcome
labels, with an informed teacher that sees the judge's input plus the outcome, explains each label it keeps, and
abstains when the outcome cannot be explained from what the judge sees. On a held-out dataset with an execution
oracle, calibration raised chance-corrected agreement (Scott's $\pi$) from 0.36 to 0.60.
On a deployed agent's outcome labels, with 25\% flipped at random to introduce noise, the teacher set aside flipped \fail{} labels at more than
twice the rate of genuine ones, cutting overall label noise from 25\% to 20\%.
Recalibrating that agent's judge using AutoRefine raised its agreement with merge/abandon outcomes on 17 independently
held-out diffs from the current workflow ($\pi$ 0.16 to 0.51); on 16 diffs from older workflows, the previous judge
stayed ahead, illustrating the need for recalibration as the agent evolves. By abstaining, the teacher not only removes noisy labels but also flags changes whose outcome the
judge's input cannot explain, telling a team when a judge needs better inputs rather than a better prompt.
\end{abstract}
\section{Introduction}
\label{sec:intro}

Coding agents now author code changes (diffs) at scale, and most of what they produce is not verifiable by a test:
refactors, performance optimizations, cleanups and open-ended fixes. In an agent's pipeline, an LLM
judge~\cite{zheng2023judging} is an increasingly common quality-assurance step for such changes: it helps decide which
diffs reach human reviewers. A judge, however, is only as good as its calibration (fitting the prompt to labels) for the task at hand: even
strong judges are unreliable on objective correctness~\cite{tan2024judgebench}, and agreement with humans varies so
much across evaluation tasks that no judge is a drop-in replacement~\cite{bavaresco2024llmjudges}. Calibrating against manually labeled datasets is resource intensive, and the labels go stale whenever the agent, the model or the code distribution moves, so in
practice many judge prompts are never validated at all. Keeping a judge trustworthy is therefore a life-cycle task:
calibration must run unattended and be rerun whenever the agent it judges evolves.

The development workflow already emits labels continuously: tests pass or fail, and changes land or are abandoned.
We expect merge and abandon outcomes to be most informative when the judge evaluates an automated system that
produces changes, and our use case is consistent with this (Sec.~\ref{sec:workflow:ma}). Outcome labels can still mislead a judge
in two ways. The outcome may be decided by information the judge never sees, such as a hidden test, or by something
other than the code, such as scope, process or a broken pipeline. A judge fitted to these labels learns to
predict noise.

We present AutoRefine, a loop that rewrites a deployable judge's prompt to reproduce outcome labels, together with
an \emph{informed teacher} that curates those labels before the loop sees them. The teacher reads exactly what the
judge will read plus the outcome. It explains each label it keeps, and when the outcome cannot be explained from
the judge's view it abstains, removing the sample instead of inventing a label. Its abstention rate doubles as a
monitoring signal: when the teacher abstains on many samples, the judge needs better inputs, not a better prompt.
We contribute:
\begin{itemize}
  \item \textbf{The loop and the abstaining, explaining teacher} (Sec.~\ref{sec:method}), used as a
  quality-assurance component on an internal agent-evaluation platform.
  \item \textbf{Calibration under trustworthy labels} (Sec.~\ref{sec:oracle}): on an execution oracle,
  calibration significantly improves a rubric judge beyond what threshold tuning achieves.
  \item \textbf{Reducing label noise by abstention, and keeping a workflow's judge current} (Sec.~\ref{sec:workflow}):
  in a controlled noise test and on a performance-optimization agent's merge/abandon labels, plus the recalibration of that agent's
  judge after the agent evolved.
\end{itemize}
\section{Method}
\label{sec:method}

\subsection{Setting}
We call a reviewable unit of code change a \emph{diff}. A deployable judge sees a fixed view $x$ of a diff: its
title, summary and code changes, plus a reference patch (Sec.~\ref{sec:oracle:results}) or a test plan
(Sec.~\ref{sec:workflow}). Each diff
carries a label $y \in \{\pass, \fail\}$ from the workflow, such as an execution oracle or the diff's terminal
review status; the label source is pluggable, so manually curated labels work the same way. The goal is a judge
prompt whose verdicts on $x$ reproduce $y$ on diffs it has not seen. The loop needs only the judge's view of an
output and an accept or reject outcome, so it applies to any agent whose outputs receive such outcomes.

\subsection{Calibration loop}
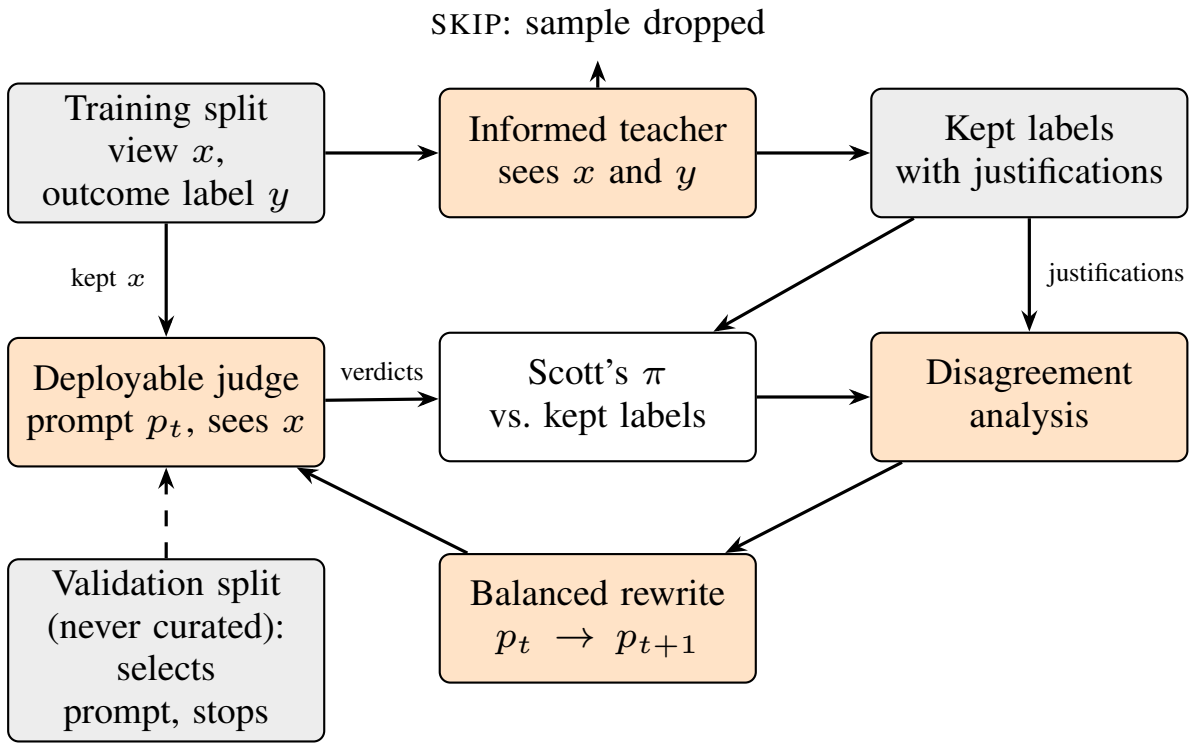
\begin{figure}[t]
\centering
\begin{tikzpicture}[font=\scriptsize,
  box/.style={draw, rounded corners=2pt, align=center, inner sep=2pt, minimum height=0.85cm, text width=1.95cm},
  llm/.style={box, fill=orange!22}, dat/.style={box, fill=gray!14},
  arr/.style={-{Stealth[length=4pt]}, semithick}]
  \node[dat] (data) {Training split\\view $x$, outcome label $y$};
  \node[llm, right=0.75cm of data] (teacher) {Informed teacher\\sees $x$ and $y$};
  \node[dat, right=0.75cm of teacher] (kept) {Kept labels\\with justifications};
  \node[llm, below=0.75cm of data] (judge) {Deployable judge\\prompt $p_t$, sees $x$};
  \node[box, below=0.75cm of teacher] (score) {Scott's $\pi$\\vs.\ kept labels};
  \node[llm, below=0.75cm of kept] (gap) {Disagreement\\analysis};
  \node[llm, below=0.6cm of score] (rewrite) {Balanced rewrite\\$p_t \rightarrow p_{t+1}$};
  \node[dat, below=0.6cm of judge] (val) {Validation split\\(never curated):\\selects prompt, stops};
  \node[above=0.18cm of teacher, font=\scriptsize] (skip) {\skipv: sample dropped};
  \draw[arr] (data) -- (teacher);
  \draw[arr] (teacher) -- (kept);
  \draw[arr, dashed] (teacher) -- (skip);
  \draw[arr] (data) -- node[left, font=\tiny] {kept $x$} (judge);
  \draw[arr] (judge) -- node[above, font=\tiny] {verdicts} (score);
  \draw[arr] (kept) -- (score);
  \draw[arr] (score) -- (gap);
  \draw[arr] (kept) -- node[right, font=\tiny] {justifications} (gap);
  \draw[arr] (gap) -- (rewrite);
  \draw[arr] (rewrite) -- (judge);
  \draw[arr, dashed] (val) -- (judge);
\end{tikzpicture}
\caption{The AutoRefine loop (orange: LLM calls; gray: data). The informed teacher curates the training labels once:
it keeps a label with a justification grounded in the judge's view, or abstains (\skipv) and drops the sample. Each
iteration runs the deployable judge on the kept samples and scores it against their labels; disagreement analysis
contrasts the judge's reasoning with the teacher's justification and proposes balanced prompt edits. An uncurated
validation split selects the final prompt.}
\label{fig:loop}
\vspace{-15pt}
\end{figure}

The loop (Fig.~\ref{fig:loop}) is an iterative optimizer over the judge's prompt. The prompt plays the role of the parameters, agreement
with the labels on the training split is the objective, disagreement analysis supplies the update direction, and a
validation split decides when to stop. Each iteration runs the current prompt on the training split and scores its
verdicts with Scott's $\pi$~\cite{scott1955pi}, which corrects raw agreement for the agreement expected by chance
from the pooled verdict and label distribution, so a judge that always predicts the majority class scores at or
below zero ($-0.08$ on our oracle test split). For every disagreement, an LLM (Claude Opus 4.7 in our runs) reads the judge's verdict and reasoning next to the
label and any teacher justification and proposes edits. The edits must be balanced, so every new reason to fail is
paired with a reason to pass, which keeps the judge from drifting toward rejecting everything. A second call applies
the edits under a length cap. For rubric judges, which emit a weighted score, the decision threshold is chosen on the
training split. A separate validation split, never curated, selects the final prompt; the loop stops at a target
agreement, when no disagreements remain, or at a maximum number of iterations (five in our runs, on training splits
of 125 to 127 diffs). The agent's
patches are recorded once and replayed, so calibration never reruns the agent under test.

\subsection{The abstaining, explaining informed teacher}
\label{sec:method:teacher}
Before calibration, the teacher reviews each training sample. In this paper it sees exactly the judge's view $x$
plus the label $y$. It has three jobs. It \emph{explains} the label, writing a justification grounded in $x$; this
is what the disagreement analysis uses to tell the rewriter why the judge was wrong. It may \emph{correct} the label
in either direction when $x$ clearly contradicts it, though in our runs it rarely did (2 of 512 merge/abandon diffs). And it \emph{abstains}, returning \skipv{} when $y$ cannot be
explained from $x$; abstained samples are dropped from calibration and from disagreement analysis. By design the
teacher prefers abstaining to a low-confidence flip, so it never invents a label it cannot support. The teacher
thus performs outcome-aware filtering and explanation, loosely analogous to distillation with privileged
information~\cite{lopezpaz2016unifying,hinton2015distillation}; reviewer comments, which our dataset builder already
collects, would add genuine side information by showing why a diff was abandoned.

A label cannot be explained from the judge's view when it is wrong, or when the view lacks the evidence for it. In
the main oracle setting (Sec.~\ref{sec:oracle:results}), both are less likely: execution outcomes make the labels
reliable, and the judge also sees the reference patch, which helps explain why a patch fails. There, we calibrate on
the labels directly.
\section{Calibration Under Trustworthy Labels}
\label{sec:oracle}

We first ask whether the loop works when the labels can be trusted, using execution outcomes as labels. This experiment serves two purposes: ground-truth labels let us validate the method before trusting
it on noisier labels, and, because test coverage is uneven, a judge calibrated on the diffs that have tests can score
similar diffs that have none.

\subsection{Setup: benchmark and splits}
The benchmark is built from landed diffs in an industrial monorepo, each with fail-to-pass tests. One agent (Claude
Opus 4.6) attempted 579 tasks; after dropping empty patches (trivial failures with no signal for a judge) we keep 417 records (353 \pass{}, 64 \fail{}). They are
split by task into training (127, rebalanced to about 70/30 so the loop sees enough failures), validation (85) and
test (85, with 13 failures); the last two keep the natural ratio of about 85/15. The other 120 records are surplus
passes left out by the rebalancing. We calibrate two agentic judges
(Claude Opus 4.8) that compare the agent's patch with the production reference patch: a rubric judge that scores
five weighted checks, and a pass/fail judge that returns a verdict. Tests are stripped from both patches, so neither
judge sees the oracle. We use these labels directly, without the teacher (Sec.~\ref{sec:method:teacher}).

\subsection{Result: calibration catches failures beyond threshold tuning}
\label{sec:oracle:results}
\begin{table}[t]
\caption{Held-out test results (85 tasks, 13 failures), before and after calibration. Each replay rescores the same
recorded patches; values are means over three replays, excluding tasks whose judge timed out or errored (about 5\% of
tasks per replay). $^\dagger$One replay.}
\label{tab:oracle}
\centering
\footnotesize
\setlength{\tabcolsep}{3pt}
\begin{tabular}{@{}lcccc@{}}
\toprule
Judge & \fail{} rec. & \fail{} prec. & \pass{} rec. & $\pi$ \\
\midrule
Rubric, uncal.\ (default thr.) & 0.50 & 0.42 & 0.89 & 0.36 \\
Rubric, uncal.\ (train thr.) & 0.69 & 0.35 & 0.80 & 0.33 \\
\textbf{Rubric, calibrated} & \textbf{0.64} & \textbf{0.68} & \textbf{0.95} & \textbf{0.60} \\
\midrule
Pass/fail, uncalibrated & 0.38 & 0.80 & 0.99 & 0.46 \\
Pass/fail, calibrated$^\dagger$ & 0.73 & 0.53 & 0.90 & 0.54 \\
\bottomrule
\end{tabular}
\vspace{-10pt}
\end{table}

Calibration raised the rubric judge from $\pi = 0.36$ to $0.60$ (Table~\ref{tab:oracle}), and the threshold does not
explain the gain. At the threshold chosen on the training split, the uncalibrated prompt reaches similar failure
recall only by flagging four times as many good patches (\pass{} recall 0.80 vs.\ 0.95), for $\pi = 0.33$; even the
best threshold picked on the test set itself gives only 0.42. Against the train-threshold baseline the paired gain is
$+0.27$ to $+0.30$ in $\pi$, with the 95\% bootstrap interval above zero and McNemar $p \le 0.013$ in every replay; we
report paired intervals rather than single point estimates, in the spirit of uncertainty-aware evaluation of AI
products~\cite{albertini2026scfc}. The uncalibrated pass/fail judge rarely failed a patch;
calibration nearly doubled its failure recall (0.38 to 0.73) at the cost of precision (0.80 to 0.53), for $\pi$
0.46 to 0.54. Paired by task against each uncalibrated replay, the calibrated judge's single replay caught 3 to 4
failures that the uncalibrated judge missed, and missed none that it caught. Rubric replays are stable
($\pi$ standard deviation 0.018).

\subsection{Result: without the reference, failures are invisible}
\label{sec:oracle:reffree}
Without the reference patch, the judge's view holds less evidence, so the teacher curates the training labels. The same loop then reached only $\pi = 0.35$ on the 85-task validation split (from $-0.08$ for the
uncalibrated starting prompt, which passed everything), catching 5 of its 13 failures; because the teacher also set
aside most training failures, the loop had few failures to learn from. To check whether the failures were visible
at all, we ran the teacher retrospectively on the validation split, which calibration never
curates. It abstained on 12 of the 13
failures and on none of the 72 passes, each time with the same reason: the patch reads as correct, and the failure
lives only in the hidden tests. Abstention thus flagged the labels the judge's view cannot explain, without the
teacher seeing the tests itself.

To test whether people can see these failures, engineers reviewed 56 teacher-curated tasks blind to the outcome and to
the teacher (25 abstained failures, 11 kept failures, 20 kept passes, drawn from the training and validation splits; two
independent rating teams, about five minutes per task). Seeing only the title, summary and patch, they judged each patch correct, incorrect or undecidable. Both
teams missed nearly all failures, whether the teacher had abstained on them (23 and 22 of 25 not called incorrect) or
kept them (10 of 11 each); across both teams, 54\% of failure ratings called the patch correct and 36\% undecidable.
They did better on passes, calling 12 and 9 of 20 correct. The judge's inputs are thus not enough to see these
failures, which is the condition the teacher's abstention flags. Sec.~\ref{sec:workflow:ma} shows that the teacher
also separates visible failures from invisible ones. For the kept failures, the teacher, reading with the outcome in hand, cited
specific defects, and 4 of 5 mechanically checkable claims held up. The two teams agreed on which outcomes they missed on 79\% of tasks
($\kappa = 0.43$).
\section{Reducing Label Noise by Abstention, and a Calibrated Judge}
\label{sec:workflow}

\subsection{Result: abstention reduces injected label noise}
\label{sec:workflow:noise}
To measure noise reduction directly, we flipped 25\% of the labels at random (fixed seed) in two settings and ran the
teacher (Claude Opus 4.8) unchanged: the performance-optimization agent's 185-diff calibration set (Sec.~\ref{sec:workflow:ma}),
where defects are often visible in the diff, and the 127-task training split of the reference-free oracle study
(Sec.~\ref{sec:oracle:reffree}), where failures are invisible from the patch. On the calibration set, the teacher set
aside 13 of the 24 landed diffs relabeled abandoned but only 16 of the 73 genuinely abandoned ones (Fisher
$p = 0.004$). Label noise fell from 24.9\% to 20.4\%; dropping the same number of labels at random within each label
leaves 24.9\% ($p < 0.001$). On the oracle split, noise fell from 25.2\% to 14.0\%, but the teacher set aside genuine
failures nearly as often as flipped ones (20 of 27 vs.\ 19 of 21), so its gain matches a rule that drops every
\fail{} label. In both settings it abstained rather than relabeled (1 relabel in 185 labels, none in 127).
It was also asymmetric: it kept most failing diffs labeled \pass{} (19 of 22 and 10 of 11). Errors that look like
success to a reader of the diff pass through. This preference is by design, set by the teacher's prompt, to prioritize
recall over precision of good diffs and can be tuned to the use case.

\subsection{Result: the teacher abstains where abandonment is not in the code}
\label{sec:workflow:ma}
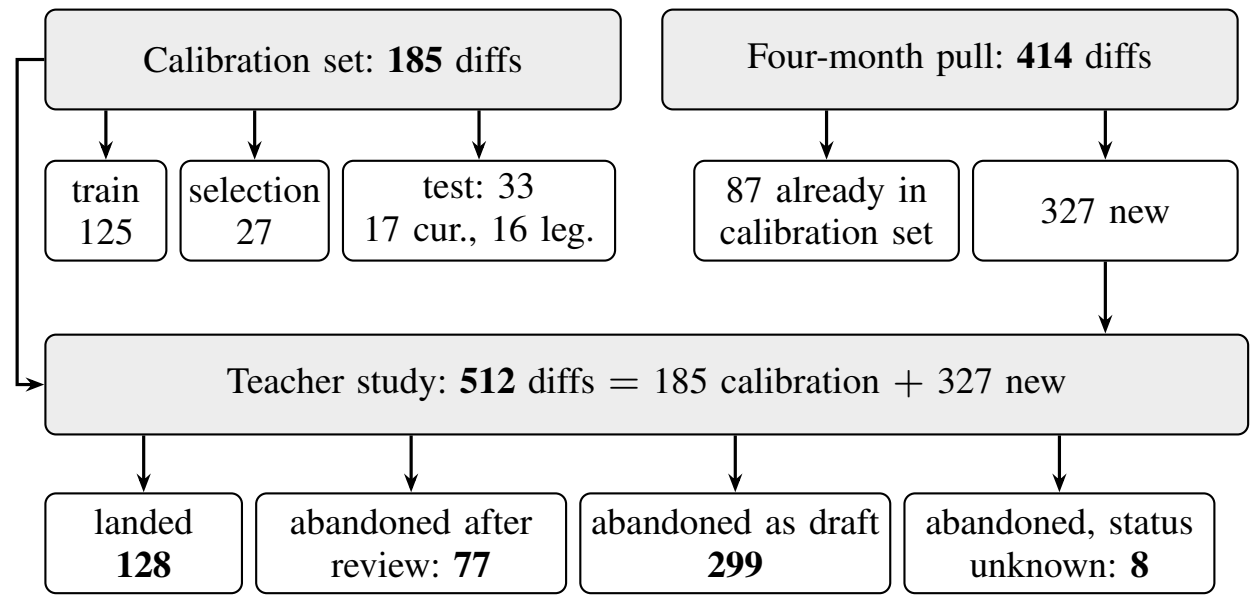
\begin{figure}[t]
\centering
\begin{tikzpicture}[font=\scriptsize,
  box/.style={draw, rounded corners=2pt, align=center, inner sep=1.5pt, minimum height=0.66cm},
  dat/.style={box, fill=gray!14}, sub/.style={box, fill=white},
  arr/.style={-{Stealth[length=3.5pt]}, semithick}]
  \node[dat, text width=3.7cm, anchor=north west] (cal) at (0,0) {Calibration set: \textbf{185} diffs};
  \node[dat, text width=3.7cm, anchor=north east] (pull) at (7.9,0) {Four-month pull: \textbf{414} diffs};
  \node[sub, minimum width=0.8cm, anchor=north west] (tr) at (0,-1.0) {train\\125};
  \node[sub, minimum width=0.95cm, right=0.08cm of tr] (sel) {selection\\27};
  \node[sub, minimum width=1.8cm, right=0.08cm of sel] (te) {test: 33\\17 cur., 16 leg.};
  \node[sub, minimum width=1.75cm, anchor=north east] (new) at (7.9,-1.0) {327 new};
  \node[sub, minimum width=1.75cm, left=0.08cm of new] (ov) {87 already in\\calibration set};
  \foreach \n in {tr,sel,te} \draw[arr] (cal.south -| \n.north) -- (\n.north);
  \foreach \n in {ov,new} \draw[arr] (pull.south -| \n.north) -- (\n.north);
  \node[dat, text width=7.85cm, anchor=north west] (all) at (0,-2.15) {Teacher study: \textbf{512} diffs $=$ 185 calibration $+$ 327 new};
  \draw[arr] (cal.west) -- ++(-0.18,0) |- (all.west);
  \draw[arr] (new.south) -- (new.south |- all.north);
  \node[sub, minimum width=1.3cm, anchor=north west] (ld) at (0,-3.2) {landed\\\textbf{128}};
  \node[sub, minimum width=2.05cm, right=0.08cm of ld] (ar) {abandoned after\\review: \textbf{77}};
  \node[sub, minimum width=2.05cm, right=0.08cm of ar] (dr) {abandoned as draft\\\textbf{299}};
  \node[sub, minimum width=2.05cm, right=0.08cm of dr] (un) {abandoned, status\\unknown: \textbf{8}};
  \foreach \n in {ld,ar,dr,un} \draw[arr] (all.south -| \n.north) -- (\n.north);
\end{tikzpicture}
\caption{Where the performance-optimization agent's diffs come from. The class-balanced calibration set supplies the splits of
Sec.~\ref{sec:workflow:recal}; adding the 327 new diffs of an uncurated four-month pull gives the 512 diffs of
Sec.~\ref{sec:workflow:ma}.}
\label{fig:data}
\vspace{-10pt}
\end{figure}

We next apply the teacher to the labels it was designed for. The performance-optimization agent in our case study submits diffs
for review; the label is the diff's terminal status (landed $=$ \pass{}, abandoned $=$ \fail{}). What ``abandoned''
means depends on the pipeline: many diffs are abandoned while still drafts, largely after internal triage by the team
operating the agent and before they are published for code review, while others are abandoned after a code reviewer
has seen them. Review metadata fetched from the code-review system shows
that most abandoned diffs were never published (Fig.~\ref{fig:data}).

\begin{figure}[t]
\centering
\includegraphics[width=\columnwidth]{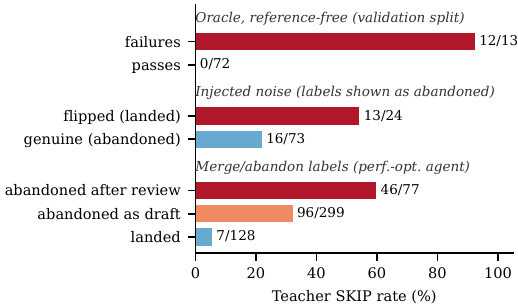}
\caption{Where the teacher abstains. On the oracle benchmark's validation split, it abstains on 12 of the 13
hidden-test failures and on none of the 72 passes; with injected noise on merge/abandon labels, it sets aside landed
diffs relabeled abandoned more than twice as often as genuinely abandoned ones; and it abstains most on diffs
abandoned after review. Labels on bars give abstentions out of $n$.}
\label{fig:abstain}
\vspace{-10pt}
\end{figure}

We ran the teacher on the 512 terminal diffs of Fig.~\ref{fig:data}. The teacher sees the title,
summary, diff and test plan plus the outcome. It abstained on 60\% of diffs abandoned after review, 32\% of drafts and
5.5\% of landed diffs (Fisher $p = 10^{-5}$ for after-review vs.\ drafts; Fig.~\ref{fig:abstain}). The pattern
holds in the 327 new, uncurated diffs (60\% vs.\ 37.5\%, $p = 0.002$) and in the class-balanced calibration set (58\% vs.\ 16\%,
$p = 0.003$), and a fresh teacher run agrees with the decisions logged during the original calibration on 112 of 125
diffs ($\kappa = 0.83$). The gap points the same way before the judge became part of the internal triage workflow (39\% vs.\ 23\% on 200
abandoned diffs, $p = 0.10$, not significant) and after
(71\% vs.\ 45\%, $p = 0.002$).

The teacher's stated reasons show why. Abstained review-abandoned diffs read as ``a clean, behavior-preserving
code-motion optimization'' with no defect visible in the diff; kept abandoned diffs cite visible problems, such as
work relocated rather than removed, noise-level gains, or a summary that contradicts its own measurements. The teacher
withholds labels where the reason for abandonment is not in the code, the same behavior as on the hidden-test failures
of Sec.~\ref{sec:oracle:reffree}. Reviewer comments support this reading. Among the diffs abandoned after review whose comments state a reason,
the comments cite a problem visible in the diff for 13 of the 19 the teacher kept but only 14 of the 38 it abstained on
(Fisher $p = 0.048$); the others cite duplicates, superseded work or a cleanup bot closing stale diffs.

\subsection{Result: recalibration raises the judge's recall}
\label{sec:workflow:recal}
The agent's initial judge ($v_1$) had been tuned on diffs from an earlier version of the agent's workflow. We
recalibrated it with AutoRefine ($v_2$, rubric judge, Claude Opus 4.8) on 125 training diffs from the calibration
set, of which the teacher set aside 19, and selected the best iteration's prompt on the 27 selection diffs
(Fig.~\ref{fig:data}); $v_2$ then replaced $v_1$ in that internal workflow. The judge first ran in shadow mode to help the team
triaging the agent's diffs publish fewer bad diffs without losing good ones, so recall on landed diffs is the
deployment metric, and a teacher that keeps \pass{} labels (Sec.~\ref{sec:workflow:noise}) protects it.
Table~\ref{tab:recal} reports the held-out slices. On 43 diffs from the then-current distribution that were not used
for training (26 of the 27 selection diffs, one lost to a $v_1$ infrastructure failure, plus 17 test diffs), $\pi$
rose by a paired $+0.31$ (95\% CI $[0.03, 0.61]$) and recall on landed diffs rose from 12/20 to 18/20, the deployment
target of 0.90; recall on abandoned diffs barely moved (11/23 to 12/23). The 17 test diffs, never used for training or
selection, form the independent test: there $\pi$ rose from 0.16 to 0.51 ($+0.35$, CI $[0.00, 0.74]$) and landed recall
from 6/8 to 8/8, a small sample whose gain matches the 43-diff one. On 16 older test diffs, $v_1$ stays ahead; because the test split mixes
two distributions, we report its current and legacy slices separately. We checked that these older diffs come from an earlier version
of the agent's pipeline: 12 of the 16 were produced by three optimization workflows that account for only 3 of the 17
current diffs.
We also checked whether $v_1$'s verdict, once visible to the triaging team, could have biased draft
abandonment: 36 of the 43 evaluation diffs (and all 16 older legacy diffs) predate $v_1$'s verdict becoming part of the workflow and carry no $v_1$ annotation, so $v_1$ cannot have
shaped their labels. On these 36 the gain is larger than on all 43 ($+0.39$, CI $[0.10, 0.70]$; landed recall 9/15 to 14/15).

\begin{table}[t]
\caption{Judge performance before ($v_1$) and after ($v_2$) recalibration, on held-out diffs. Each cell gives $v_1 \rightarrow v_2$. Landed (abandoned) recall: landed
diffs judged \pass{} (abandoned diffs judged \fail{}); landed recall is the deployment metric (target 0.90). Only the test slices were never used for selection.
$^\diamond$Older diffs from the earlier pipeline $v_1$ was tuned on, assigned to the test split in advance; $v_2$'s drop here is consistent with
distribution shift, and is why we recalibrate as the agent evolves.}
\label{tab:recal}
\centering
\footnotesize
\setlength{\tabcolsep}{2pt}
\begin{tabular}{@{}lcccc@{}}
\toprule
Slice & $n$ & $\pi$ & Landed & Abandoned \\
\midrule
Current (sel.+test) & 43 & $0.07 \rightarrow \mathbf{0.38}$ & $12/20 \rightarrow \mathbf{18/20}$ & $11/23 \rightarrow \mathbf{12/23}$ \\
\quad indep.\ test & 17 & $0.16 \rightarrow \mathbf{0.51}$ & $6/8 \rightarrow \mathbf{8/8}$ & $4/9 \rightarrow \mathbf{5/9}$ \\
\quad before $v_1$ & 36 & $0.05 \rightarrow \mathbf{0.44}$ & $9/15 \rightarrow \mathbf{14/15}$ & $10/21 \rightarrow \mathbf{12/21}$ \\
Legacy test$^\diamond$ & 16 & $\mathbf{0.50} \rightarrow -0.02$ & $\mathbf{6/8} \rightarrow 5/8$ & $\mathbf{6/8} \rightarrow 3/8$ \\
\bottomrule
\end{tabular}
\vspace{-10pt}
\end{table}
\section{Related Work}
\label{sec:related}

\emph{LLM judges.} LLM-as-a-judge was validated against human preferences by Zheng et al.~\cite{zheng2023judging};
later work documents biases~\cite{ye2024justice}, weak performance on objective
correctness~\cite{tan2024judgebench}, task-dependent agreement with humans~\cite{bavaresco2024llmjudges}, and
pitfalls in how judges are validated~\cite{guerdan2025validating}.

\emph{Prompt and judge optimization.} Prompts can be optimized without weight
updates~\cite{yang2024opro,khattab2024dspy,yuksekgonul2025textgrad}. Closest to our work,
AutoCalibrate~\cite{liu2023autocalibrate} refines an LLM evaluator's criteria and EvalGen~\cite{shankar2024evalgen}
aligns evaluators with user grades, both against human labels. AutoRefine instead takes supervision from workflow
outcomes and adds a teacher that explains or withholds each label; the optimizer itself is interchangeable.

\emph{Expert validation and uncertainty in AI engineering.} Prior work motivates domain-expert
oversight~\cite{gren2026evf} and uncertainty-aware evaluation from stratified samples~\cite{albertini2026scfc}; an independent,
representative audit of production labels remains future work.

\emph{Weak supervision.} Denoising labels from noisy sources is the core of weak
supervision~\cite{ratner2017snorkel}, which combines many noisy labeling sources rather than
one teacher.

\emph{Oracles and SE for ML.} For changes without tests, the judge acts as the oracle in the broad sense of the oracle
problem~\cite{barr2015oracle}, and keeping such ML components correct as data and systems change is a core concern of
software engineering for ML~\cite{amershi2019se4ml}.

\emph{Outcomes in software engineering.} Pull-request acceptance and merged vs.\ abandoned changes have long been
prediction targets~\cite{gousios2014pullbased,soares2015acceptance,islam2022merged}. There, the outcome is the
target, and acceptance reflects many factors besides technical quality; here the outcome is a noisy label for
calibrating a judge.

\section{Lessons, Threats to Validity and Conclusion}
\label{sec:conclusion}

\emph{Lessons for practitioners.} (i)~Recalibrate whenever the agent or its workflow changes: $v_2$ wins on current
diffs, while $v_1$ still wins on the older diffs it was tuned on (Sec.~\ref{sec:workflow:recal}). (ii)~Use
merge/abandon labels for judges of agent-written diffs, and check what a label means at each pipeline stage (a
discarded draft is not a reviewer's rejection); for human-written diffs, abandonment more often reflects scope
or process, which the teacher can only set aside. (iii)~Ask reviewers to say why they abandon a diff: a comment cannot
make the diff judgeable from what the judge sees, but it would sharpen the teacher's justification, which the disagreement
analysis turns into new rules for the judge's prompt.

\emph{Threats to validity.} \emph{Construct:} workflow outcomes stand in for code quality, and abstention can also
drop useful samples, such as genuine failures that look clean. \emph{Internal:} judge and teacher share a model family,
which risks self-preference~\cite{ye2024justice}; in our runs the teacher caught correct patches labeled \fail{} but not
failing patches labeled \pass{}, and use cases that need the opposite should tune its prompt. \emph{External:} all
studies come from one source, and the merge/abandon study from one agent in use. \emph{Conclusion validity:} samples
are small, each calibration loop ran once per setting, and the human review covers 56 tasks with moderate agreement
between the two rating teams.

\emph{Conclusion.} AutoRefine calibrates a code judge from labels the workflow already produces, with a
teacher that explains the labels it keeps and abstains on the ones it cannot explain. Under trustworthy labels,
calibration raised a rubric judge's $\pi$ from 0.36 to 0.60, beyond what threshold tuning achieves. On a performance-optimization agent, recalibration raised the judge's $\pi$ from 0.16 to 0.51 on an
independent held-out test. The teacher
keeps labels it cannot explain out of calibration.

\bibliographystyle{IEEEtran}
\bibliography{IEEEabrv,paper}

\end{document}